%% file: neurips_2025.tex
\documentclass{article}

\usepackage[creativeai, final]{neurips_2025}
\usepackage{graphicx}
\usepackage{float}
\usepackage{amsmath}
\usepackage{natbib}
\usepackage[utf8]{inputenc} % allow utf-8 input
\usepackage[T1]{fontenc}    % use 8-bit T1 fonts
\usepackage{hyperref}       % hyperlinks
\usepackage{url}            % simple URL typesetting
\usepackage{booktabs}       % professional-quality tables
\usepackage{amsfonts}       % blackboard math symbols
\usepackage{nicefrac}       % compact symbols for 1/2, etc.
\usepackage{microtype}      % microtypography
\usepackage{xcolor}         % colors

\usepackage{booktabs}
\usepackage{longtable}
\usepackage{csvsimple}

\title{Scene-Aware Urban Design: A Human–AI Recommendation Framework Using Co-Occurrence Embeddings and Vision-Language Models
}
\author{%
  Rodrigo A. Gallardo \\
  Department of Architecture \\
  Department of EECS \\
  Massachusetts Institute of Technology \\
  Cambridge, MA 02139 \\
  \texttt{ragallar@mit.edu}
  \And
  Oz Fishman \\
  Department of Architecture \\
  Massachusetts Institute of Technology \\
  Cambridge, MA 02139 \\
  \texttt{oof0928@mit.edu}
  \And
  Alexander Htet Kyaw \\
  Department of Architecture \\
  Department of EECS \\
  Massachusetts Institute of Technology \\
  Cambridge, MA 02139 \\
  \texttt{alexkyaw@mit.edu}
}

\begin{document}
\raggedbottom

\maketitle

\begin{abstract}
This paper introduces a human-in-the-loop computer vision framework that uses generative AI to propose micro-scale design interventions in public space and support more continuous, local participation. Using Grounding DINO and a curated subset of the ADE20K dataset as a proxy for the urban built environment, the system detects urban objects and builds co-occurrence embeddings that reveal common spatial configurations. From this analysis, the user receives five statistically likely complements to a chosen anchor object. A vision language model then reasons over the scene image and the selected pair to suggest a third object that completes a more complex urban tactic. The workflow keeps people in control of selection and refinement and aims to move beyond top-down master planning by grounding choices in everyday patterns and lived experience.
\end{abstract}

\section{Introduction}
The quality of public urban space is shaped not only by large-scale master planning but also by small, everyday interventions: a bench beneath a tree, a bike rack near a storefront, or a shade structure in a plaza. The results of these micro-scale decisions are core to the lived experience of the city \cite{koolhaas_whatever_1995}. For residents, the ability to make such changes to their surroundings is constrained by regulation, resource access, or a lack of design knowledge \cite{velasco_housing_2018}. In their absence, municipalities’ design of these spaces is often generic or devoid of the spatial knowledge held by residents who use them daily. Our work would help residents make situated recommendations for the improvement of their surroundings, anchored on specific objects, and in a decision environment that is constructed from AI-generated suggestions of which additional objects would most likely create a more successful urban environment \cite{kindberg_people_2000, agre_surveillance_1994}. 

Rather than automating design, this hybrid model uses machine learning to surface possibilities to allow humans to provide interpretation, judgment, and intent. This collaboration raises timely questions: How can non-human agents help us to create decision environments that enable more active public participation? What kinds of local knowledge or spatial rituals can be preserved as machine-generated suggestions enter civic processes? And how might systems like this redistribute design agency, enabling new forms of engagement with the built environment? \cite{busquets_defining_2007} 

\section{State of the Art}

Tangible user interfaces that integrate “physical” manipulation with real-time simulation allow residents and planners alike to test zoning and land-use scenarios through accessible, hands-on interfaces that link directly to computational simulations \cite{noyman_urban_2022}. This foregrounding of spatial cause-and-effect democratizes access to the many decisions regarding (and the implications of) involved in macro-level interventions (such as neighborhood density, mobility networks, or district-wide zoning), but offers limited usability for the everyday, object-scale decisions that define how individuals inhabit public space, especially at scale. Emerging developments in extended reality and computer vision have enabled a new mode of interaction with the built environment, treating physical objects as responsive, queryable agents \cite{kindberg_people_2000}. Through real-time semantic segmentation and large language models, such systems allow users to direct prompts at material elements within their surroundings, effectively enabling interactions through augmented interfaces \cite{liu_grounding_2024, dogan_augmented_2024, radford_learning_2021}. Recent research has also demonstrated the use of vision-language models to integrate real-world data and sustainability metrics into design pipelines. \cite{gupta_insights_2025} These interaction paradigms reframe urban elements as digital endpoints for dialog and design. 

\section{Methods}

\begin{figure}[H]
    \centering
    \includegraphics[width=1\linewidth]{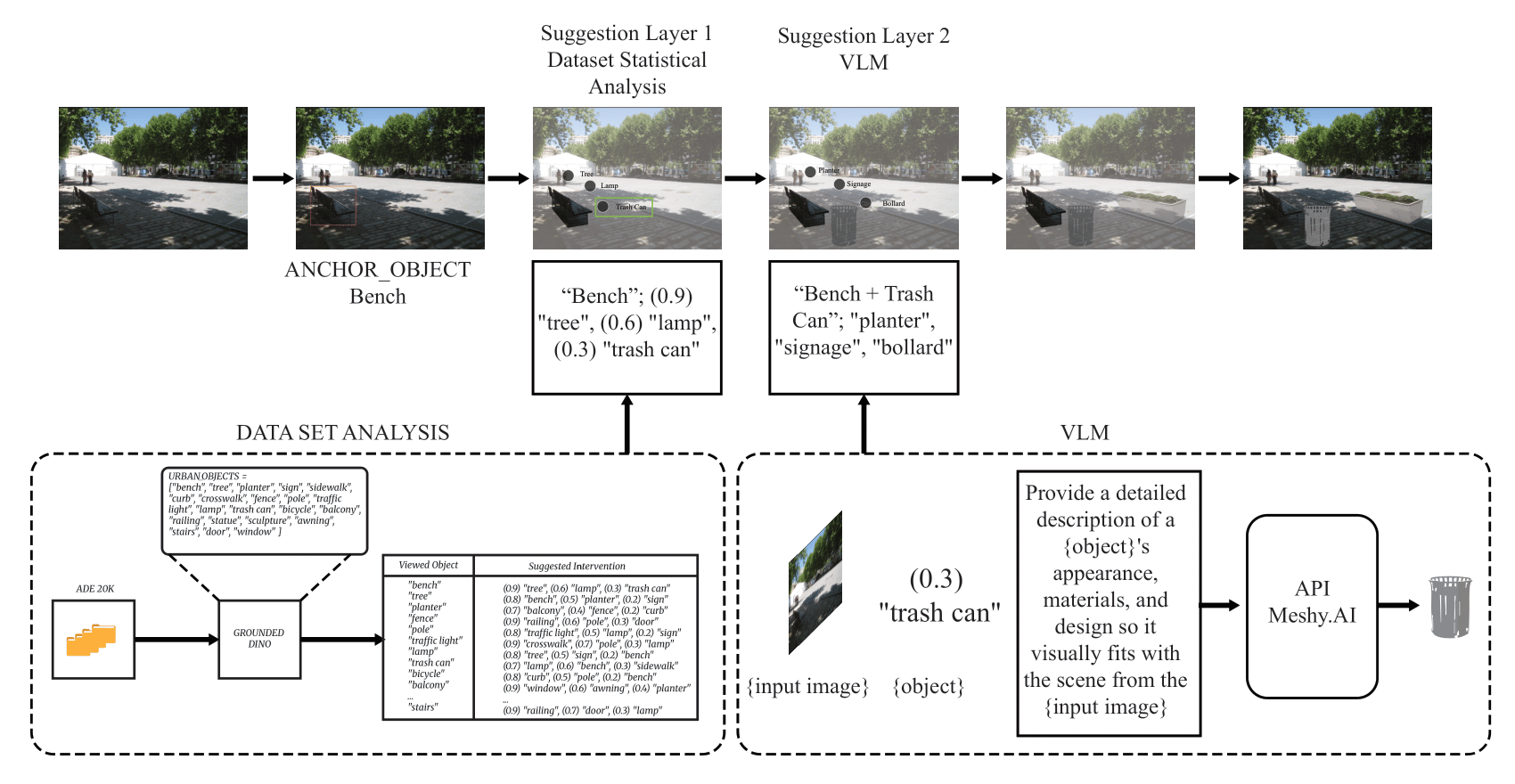}
    \caption{Overview Diagram}
    \label{fig:placeholder}
\end{figure}

The proposed pipeline runs lightweight background object detection as scenes are processed. When a scene meets the criteria for a potential intervention, an anchor object is specified for that scene by the user. Once the anchor is set, the system initiates a two-branch decision process.

In the statistical branch, the system retrieves a short list of urban objects that most frequently co-occur with the anchor in socially active contexts, based on prior co-occurrence analysis using the ADE20K dataset. The user selects one of these statistically suggested co-occurring objects.

In the semantic branch, the selected anchor and co-occurring object pair is passed to a vision-language model, which proposes five additional objects that could be added to the scene using contextual and spatial cues. (see Appendix for the full system prompt).

This staged selection process, beginning with an anchor object chosen by the user and followed by a co-occurring object suggested by the system, ensures that the resulting set of objects is grounded in real-world spatial relationships. The objective is to enable the creation of urban tactics composed of multiple interacting objects.

Once the five additional recommendations are generated, 3D models are produced for each object to give the user a visual preview. To ensure these models are visually integrated with the existing scene, we employ a vision language model (VLM). Given an image of the scene, the model is prompted to “Provide a detailed description of an {object}’s appearance, materials, and design that could visually fit with the scene from the input image.” The output description is then used as input to a text-to-3D generation model, producing scene-consistent 3D assets that the user can select and situate in the urban space.

In its current stage, user participation occurs at two levels: (1) the user selects an anchor object and chooses among system-suggested co-occurring elements, and (2) the user may iteratively accept, reject, or re-prompt VLM-generated recommendations before committing them to the scene. This iterative loop preserves user intent, ensuring that each regenerated scene remains grounded in human selection and interpretation rather than autonomous replacement.

\subsection{Dataset Analysis}

\begin{figure}[H]
    \centering
    \includegraphics[width=1\linewidth]{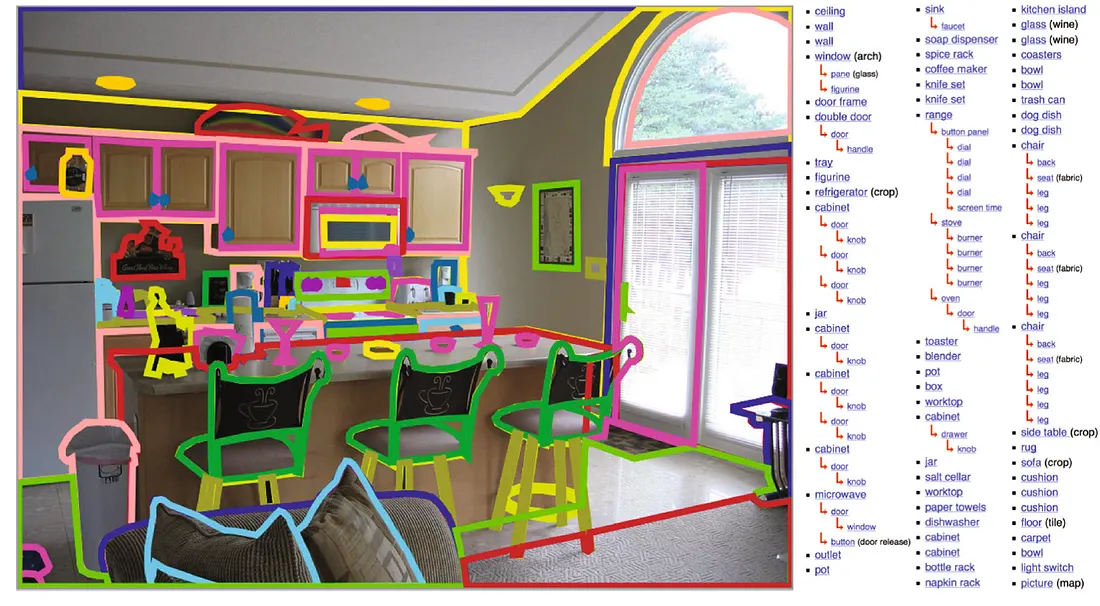}
    \caption{ADE20K image with listed objects and their parts}
    \label{fig:placeholder}
\end{figure}

The framework’s object analysis relies on running the Grounding DINO model ADE20K dataset.  ADE20K offers over 20,000 labeled images spanning indoor and outdoor environments, with dense semantic segmentation and scene categorization \cite{zhou_scene_2017}. Its breadth makes it one of the few large-scale datasets with labeled public space scenes that include furniture, vegetation, signage, and infrastructure.

Grounding DINO was selected for its open-vocabulary, zero-shot detection capabilities \cite{liu_grounding_2024}. However, the model’s reliance on text prompts introduces potential inconsistency in detection quality, especially in visually cluttered or low-resolution images. Preliminary testing also revealed false positives in highly textured scenes (e.g., planters confused with trash cans), highlighting the importance of visual post-processing and vocabulary refinement.

Specifically, scenes are only included if Grounding DINO detects five or more people with confidence above a specified threshold. This heuristic approximates scenes with higher levels of public activity, based on the assumption that activated space is correlated with human presence. While this filter does not guarantee spatial proximity or engagement, it serves as a practical proxy for identifying images where the co, occurrence of urban objects may relate to actual use and behavior.

This two-stage filtering process, first by scene category, then by population threshold, produces a set of approximately 900 images, each annotated with object detection results. Object detection is performed using Grounding DINO model. This approach allows for the detection of urban elements beyond closed, class labels, which is essential for analyzing micro-scale features often absent from conventional object detection datasets.

\subsection{Co-Occurrence Aggregation and Embedding}

\begin{figure}[H]
    \centering
    \includegraphics[width=1\linewidth]{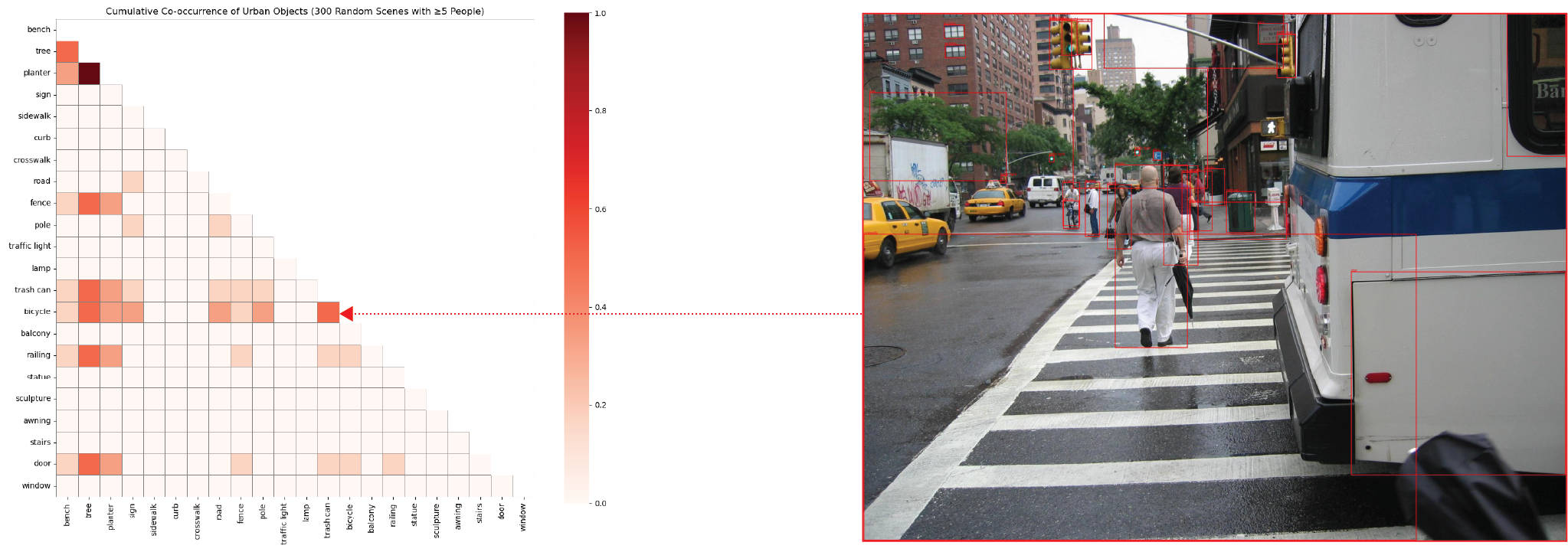}
    \caption{900 Image Co-Occurrence Matrix}
    \label{fig:placeholder}
\end{figure}

Once urban objects are detected in each filtered scene, the system constructs a pairwise co-occurrence matrix to capture which elements frequently appear together. For every qualifying image, all unordered pairs of detected urban objects are recorded in a symmetric matrix indexed by object class. If a scene includes a bench, a tree, and a trash can, for example, the matrix entries for (bench, tree), (bench, trash can), and (tree, trash can) are incremented. This aggregation process is repeated across the dataset to produce a global co-occurrence matrix spanning all scenes of interest.
{
\[
P(o_j \mid o_i) = \frac{\text{count}(o_i \wedge o_j)}{\text{count}(o_i)}
\]
}
The raw counts in the matrix are then normalized to create co-occurrence probability vectors for each object. Each row of the matrix is divided by its sum, yielding a vector that reflects the likelihood of other objects appearing in a scene given the presence of a particular object. These vectors function as contextual embeddings, not learned via backpropagation, but constructed empirically from real, world spatial data.
{
\[
o = \big[ P(o_1 \mid o),\, P(o_2 \mid o),\, \dots,\, P(o_n \mid o) \big]
\]
}
\subsection{Vision Language Model}

Vision-language models can read style, materials, and spatial cues directly from images. After the user picks an item from the co-occurrence list, the model receives three inputs: the full scene image, tight crops and normalized boxes for the anchor and the selected co-occurring object, and a short scene summary with scene type, a five-color palette, dominant materials, and a coarse near–mid–far depth sketch.

The model returns five candidates in a compact schema: object type, key materials and finish, approximate size, color hints, simple placement guidance relative to the anchor or pair, and a one-sentence justification. This format is short and easy to parse, while still carrying the information needed for the next stage. To keep suggestions practical, we filter out items that the current scene cannot support. For example, a crosswalk is often associated with a bench, but it is removed when the image does not show a street edge or an intersection. Similar feasibility checks look at ground type, clearance, and obvious access constraints before ranking the final five.

\subsection{Mesh Generation}

Off-the-shelf text-to-3D models can produce meshes that match a target style when given a clear, structured prompt. In our workflow, the VLM output is converted into a concise brief and sent to the Meshy API \cite{noauthor_api_2025} for a first-pass mesh. The system is model agnostic, so other generators can be swapped in and benchmarked for speed, geometry fidelity, UV coverage, and texture quality.

We normalize scale to real-world units, set the pivot at ground level, and generate a low-poly LOD for AR. If a mesh fails, we regenerate with a tightened prompt. 

\section{Results}

\begin{figure}[H]
    \centering
    \includegraphics[width=1\linewidth]{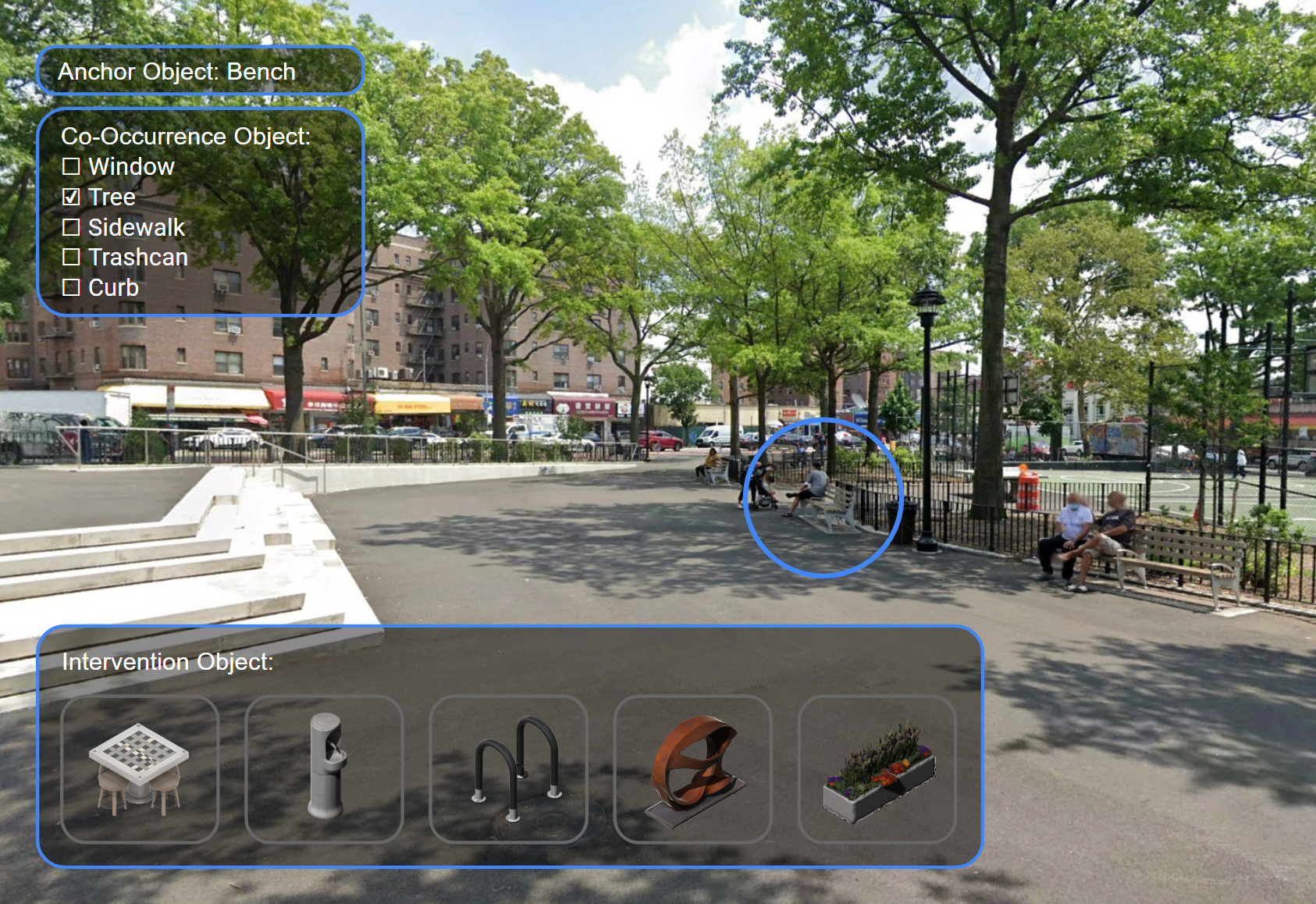}
    \caption{Pilot Interface in Urban Scene}
    \label{fig:placeholder}
\end{figure}

A pilot version of the interface was created to demonstrate the two steps involved in completing an urban tactic, using a bench in the image as the anchor object. A similar interface could be implemented in an AR environment with a headset, giving the user the ability to place and adjust models in 3D space.

Across varied urban scenes, the pipeline produced placeable suggestions. In Layer 1, the co-occurrence embedding provided in Appendix A.1 is used to provide object pairs that can complements to the anchor object, such as benches with trees, trash cans, signs, and lamp posts. We also tested the VLM with various prominent anchor objects and various co-occurrence pairings with VLM generated recommendations, as detailed in Appendix A.3. 

Examples from four selected scenes with different prominent anchor objects are shown in Appendix A.2, where the VLM produced contextually specific suggestions such as bus stops, kiosks, and wayfinding signage. These are the kind of items that a predefined co-occurrence list alone would be not be able to recommend. The ability to use a VLM in this step is critical because it can combine visual context, spatial cues, and object semantics to generate recommendations that go beyond statistical pairings, adapting to the specific scene and producing more functionally relevant, locally appropriate, and visually coherent suggestions.

Failure modes included weak detections in cluttered scenes, occasional near-duplicate or policy-sensitive suggestions in Layer 2, and certain object generation errors from the text-to-3D stage. In particular, Meshy AI sometimes failed to capture all prompt details. For example, chess table patterns were rendered inaccurately, and a drinking fountain described with a lower basin did not reproduce that feature correctly (Figure 5). These limitations highlight the need for a tighter vocabulary, region-aware guidance, stronger negative examples, and improvements in the downstream 3D generation stage.

\begin{figure}[H]
    \centering
    \includegraphics[width=0.9
\linewidth]{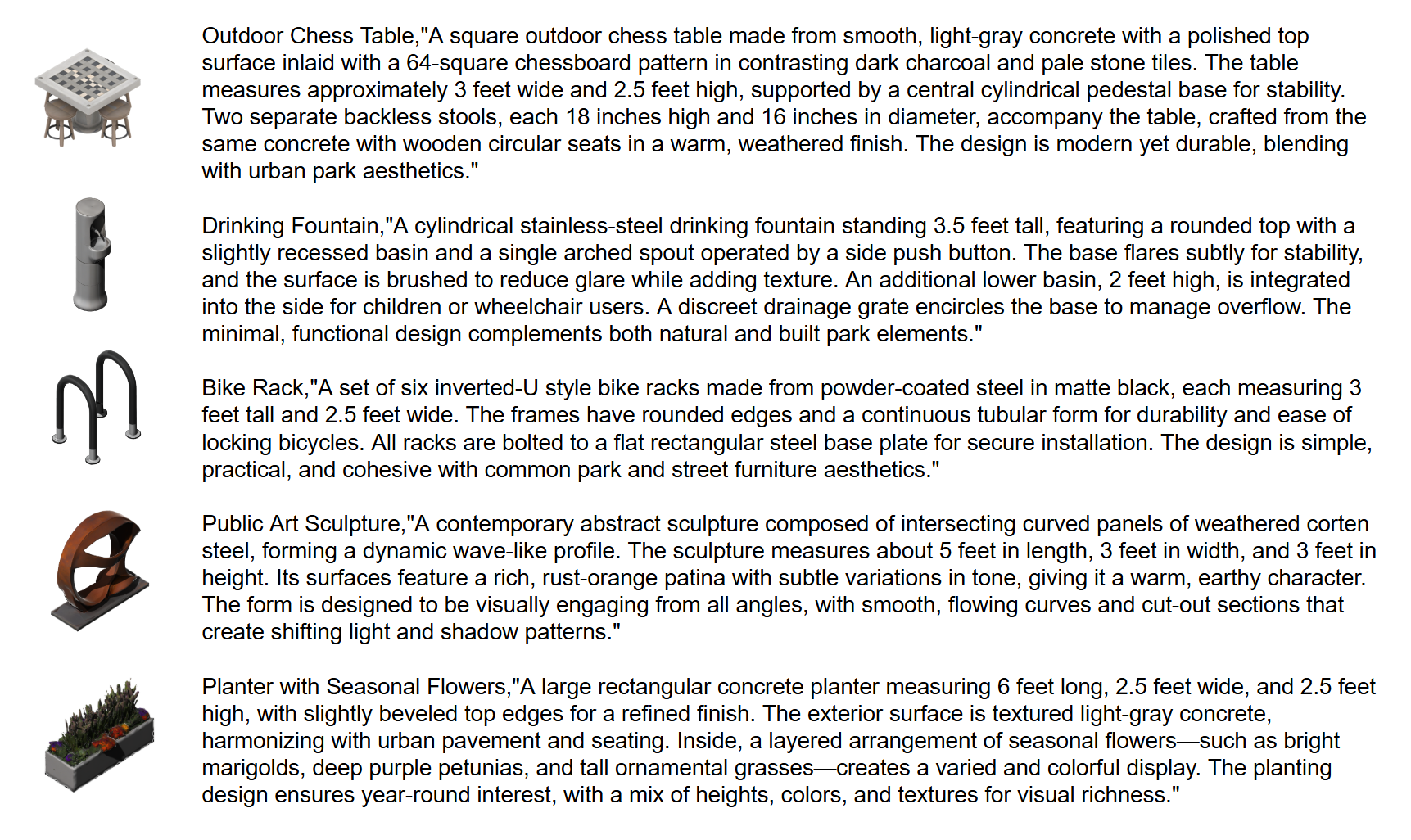}
    \caption{Mesh generated using VLM generted recommendation and discription}
    \label{fig:placeholder}
\end{figure}

\section{Limitations and Future Work}

The current system estimates object co-occurrence based on pixel distance in 2D images, which does not capture true 3D spatial relationships and may reduce precision when proposing site-specific interventions. In addition, the dataset is biased toward certain geographic and cultural contexts, limiting generalization. Future directions include integrating 3D scene understanding from RGB-D or multi-view video, adapting the interface for continuous use on AR headsets, expanding datasets for greater diversity, and incorporating post-deployment feedback to refine suggestions. There is also potential to link the system with civic reporting platforms \cite{noauthor_snap_2025}, allowing proposed interventions to be shared with local authorities and, over time, letting real-world outcomes inform future recommendations.

Vision–language models (VLMs) face several limitations when applied to recommendation tasks in urban design contexts. They may struggle to account for nuanced, location-specific social, cultural, or legal constraints, leading to suggestions that, while visually plausible, may be socially inappropriate or non-compliant with local regulations. VLMs also often lack access to critical non-visual factors such as spatial measurements, traffic patterns, and accessibility requirements, which are essential for determining the practical suitability of an object in a given location. Furthermore, biases in training data, particularly if it is dominated by imagery from specific regions such as North America or Europe can result in recommendations that overlook local architectural styles, materials, and street usage patterns in other regions. Finally, VLMs cannot directly assess installation feasibility, meaning they are unable to consider site-specific constraints such as terrain, underground utilities, or budget.

Future iterations will ideally include a participatory evaluation, where users can directly edit proposed scenes. This will allow measurement of how the AI’s recommendations influence, align with, or diverge from user intent, helping to refine regeneration fidelity and interpretability.

\section{Conclusion}

This framework is both a technical system and a design argument. As a system, it operationalizes co-occurrence embeddings and semantic reasoning within a human-in-the-loop AR interface, enabling residents to propose micro-scale interventions grounded in statistical spatial patterns and their own situated knowledge. As an argument, it contends that the city’s smallest parts (benches, trees, lights) are as consequential to urban life as its master plans, yet remain largely inaccessible to participatory design at scale. By placing computational intelligence in the immediacy of the site, the framework transforms the urban environment itself into a living model, where design emerges from dialogue between people, objects, and algorithms. In doing so, it reframes AI as a collaborator in everyday spatial authorship, embedding lived experience directly into civic processes and advancing a model of local spatial intervention.

\newpage
\bibliography{NEURIPSref}

%%%%%%%%%%%%%%%%%%%%%%%%%%%%%%%%%%%%%%%%%%%%%%%%%%%%%%%%%%%%
\newpage
\appendix

\section{Technical Appendix}

\subsection{Co-Occurence Matrices}

\begin{figure}[H]
    \centering
    \includegraphics[width=0.45\linewidth]{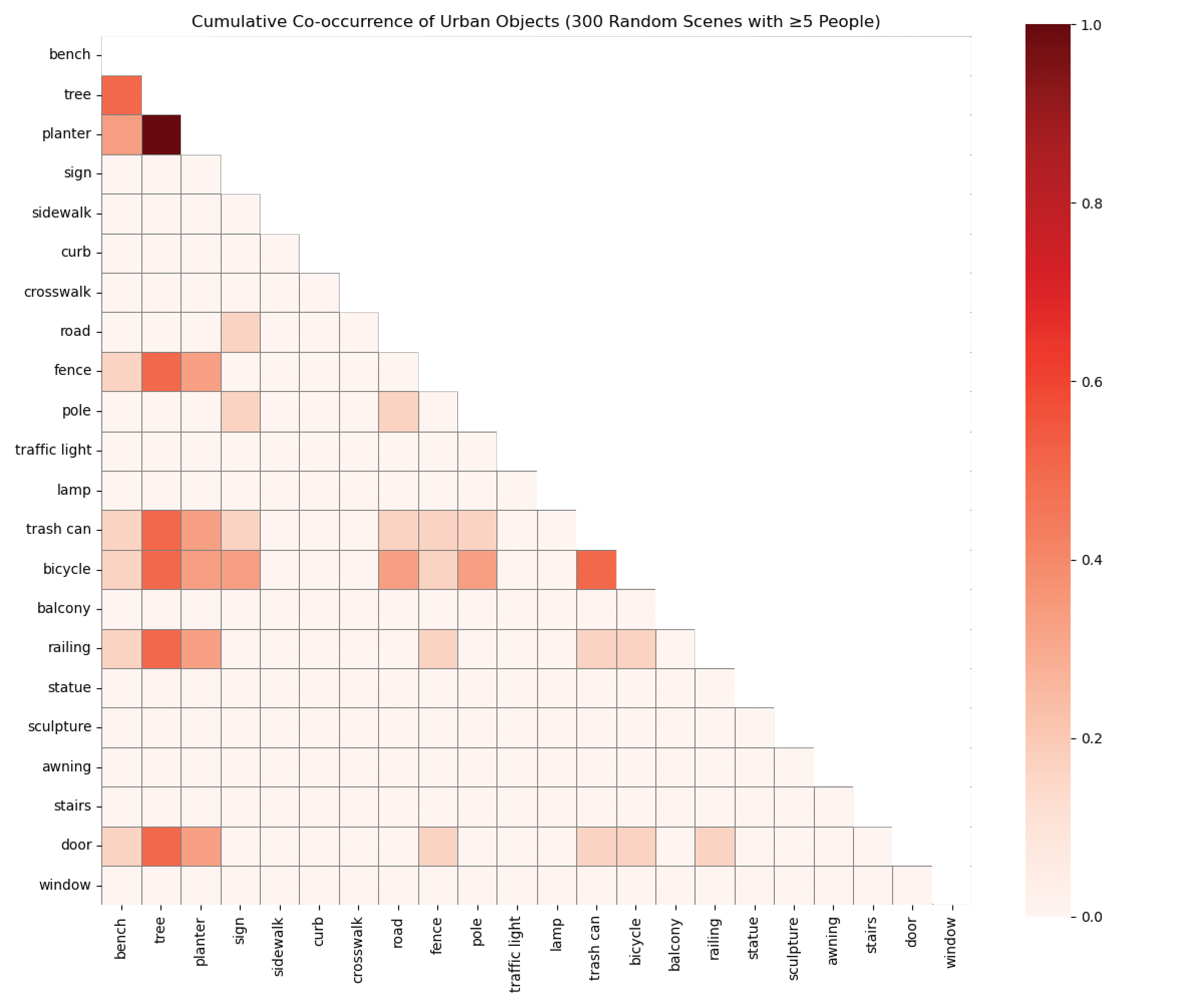}\\
    \includegraphics[width=0.45\linewidth]{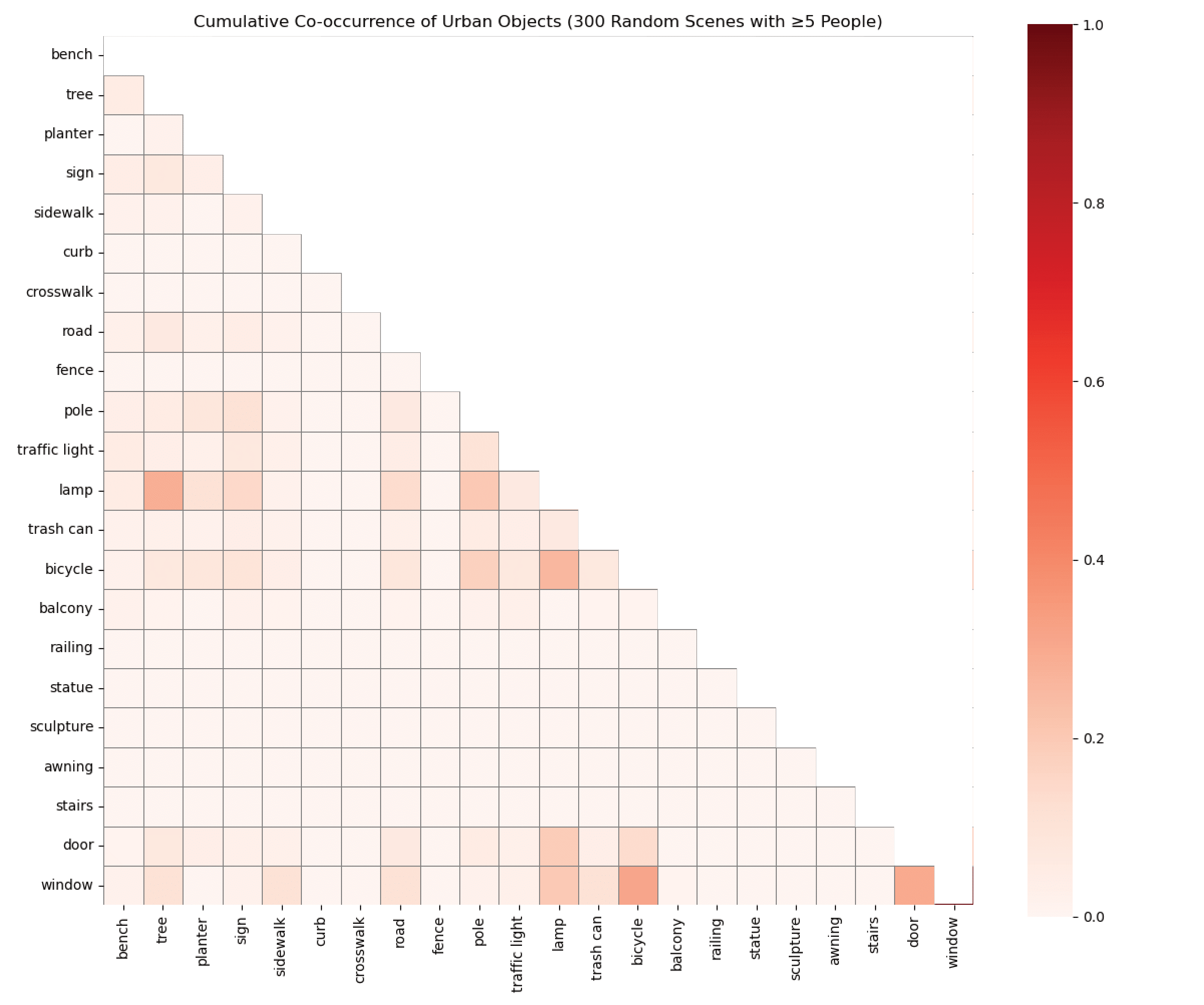}\\
    \includegraphics[width=0.45\linewidth]{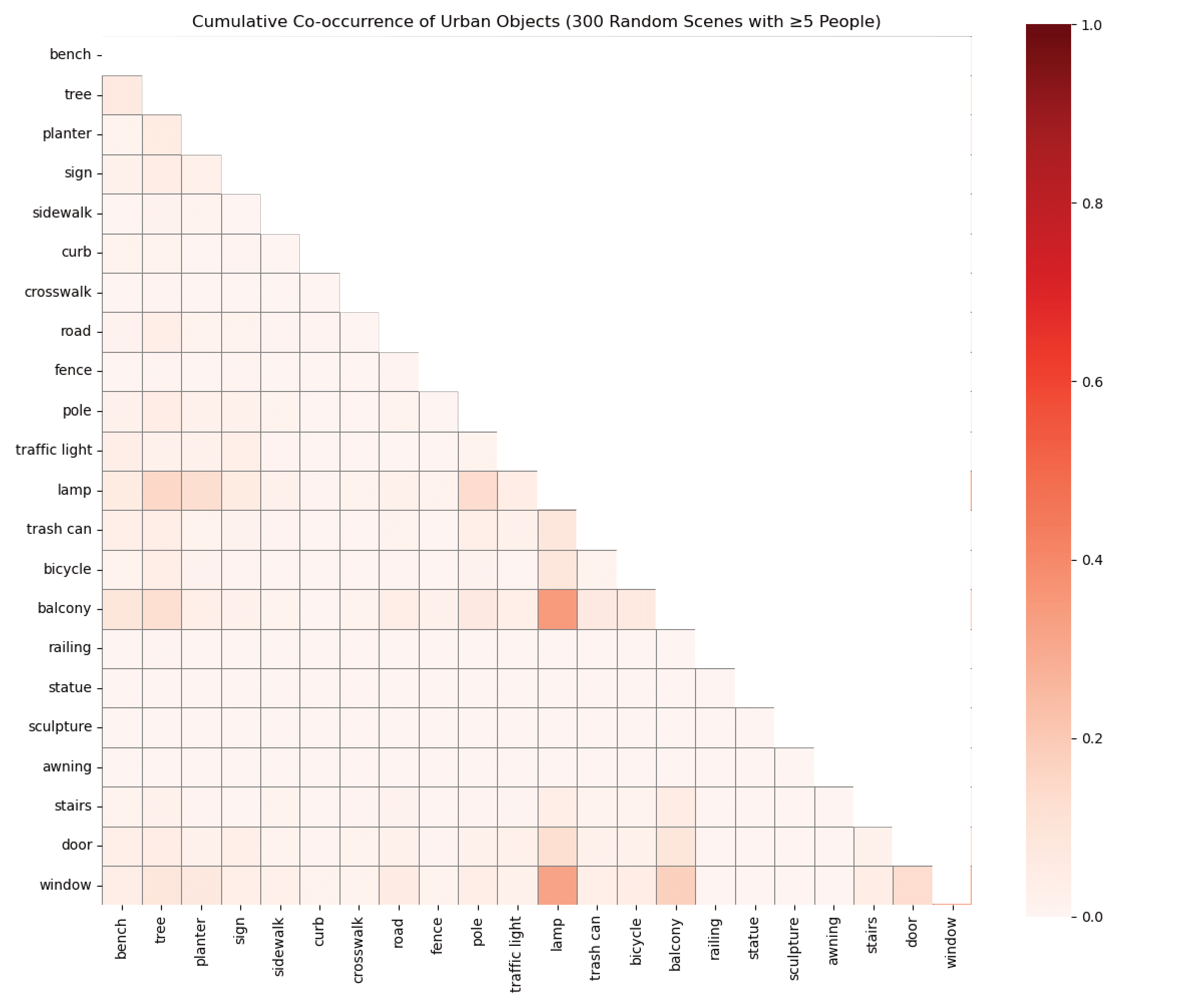}\\
    \caption{Co Ocurrence Matrices per sub folder}
    \label{fig:three_images}
\end{figure}

The system was tested on a curated subset of the ADE20K dataset, filtered to include approximately 300 urban scenes with high pedestrian density. 
For each scene, urban object detections were processed using Grounding DINO, and co, occurrence statistics were aggregated into a global matrix across all scenes of interest.
The resulting matrix revealed clear and intuitive spatial associations among frequently detected objects.

For example, the object class “bench” co, occurred most often with “tree,” “trash can,” “sign,” and “lamp post.” Similarly, “planter” frequently appeared with “fence,” “balcony,” and “sidewalk.” These pairings align with common public space configurations and validate the hypothesis that co, occurrence vectors can reflect real, world design patterns.

\subsection{Embeddings}

% Table test
\begin{table}[ht]
\centering
\caption{Top 5 Co-Occurrence Embeddings}
\vspace{0.5em} % optional small space between caption and table
\begin{center}
\resizebox{\textwidth}{!}{%
\begin{tabular}{llllll}
\toprule
\textbf{Anchor Object} & \textbf{Embedding 1} & \textbf{Embedding 2} & \textbf{Embedding 3} & \textbf{Embedding 4} & \textbf{Embedding 5} \\
\midrule
bench & window & tree & sign & traffic light & crosswalk \\
tree & traffic light & window & sidewalk & door & planter \\
planter & tree & sidewalk & window & balcony & traffic light \\
sign & traffic light & window & crosswalk & tree & sidewalk \\
sidewalk & window & traffic light & tree & planter & sign \\
curb & window & sign & traffic light & sidewalk & crosswalk \\
crosswalk & traffic light & window & sign & tree & sidewalk \\
fence & window & sidewalk & tree & planter & traffic light \\
pole & window & traffic light & tree & sign & crosswalk \\
traffic light & sign & window & tree & crosswalk & sidewalk \\
lamp & window & door & tree & stairs & sidewalk \\
trash can & window & tree & traffic light & sign & door \\
bicycle & window & traffic light & pole & fence & sidewalk \\
balcony & planter & tree & sidewalk & fence & door \\
railing & window & pole & bicycle & fence & sidewalk \\
stairs & traffic light & window & sidewalk & tree & door \\
door & tree & window & traffic light & sidewalk & crosswalk \\
window & traffic light & sidewalk & pole & fence & bicycle \\
\bottomrule
\end{tabular}%
}
\end{center}
\label{tab:top5}
\end{table}

% Table test

\subsection{System Prompt}

User Prompt
You are given an input image of an urban scene along with an anchor object and a co-occurrence object. Based on the scene in the image, propose five options for a third object that would plausibly fit into the scene in a socially active public space.

For each object, provide:

Object – The object’s name.

Description – A detailed paragraph that can be directly used for text-to-3D generation. This description must specify the object’s appearance, materials, approximate scale, color palette that matches the scene, style cues, likely placement relative to the anchor and co-occurrence objects, and functional details such as geometry, key components, and reasonable dimensions. Ground every detail in the visual context of the provided image so that the 3D model will integrate naturally into the scene. However, do not specify where the object is supposed to be placed in the scene. It should be an independent object without a background. Avoid brand names, unsafe elements, or features that block primary circulation. 

Only output the result as a CSV file with exactly two columns: Object and Description. Do not include any other commentary or formatting.

Inputs:

Scene image: {filename}

Anchor object: {object1} 

Co-occurrence object: {object2}

\subsection{Additional Examples and Scene Outputs from VLM-Generated Recommendations and Descriptions for 3D Generative AI.}

\begin{figure} [H]
    \centering
    \includegraphics[width=1\linewidth]{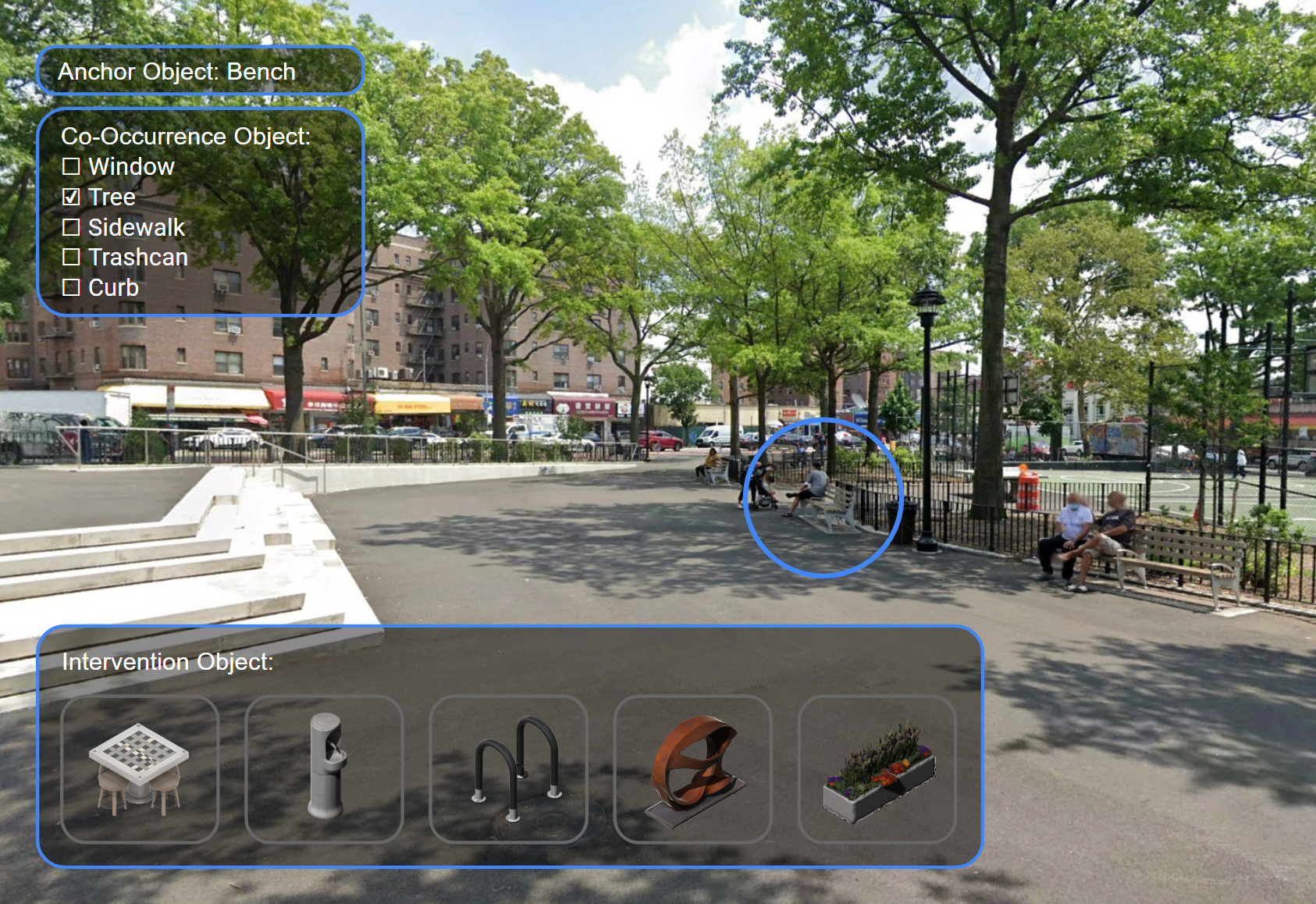}
    \caption{Pilot Interface in Urban Scene 2}
    \label{fig:placeholder}
\end{figure}

Outdoor Chess Table,"A square outdoor chess table made from smooth, light-gray concrete with a polished top surface inlaid with a 64-square chessboard pattern in contrasting dark charcoal and pale stone tiles. The table measures approximately 3 feet wide and 2.5 feet high, supported by a central cylindrical pedestal base for stability. Two separate backless stools, each 18 inches high and 16 inches in diameter, accompany the table, crafted from the same concrete with wooden circular seats in a warm, weathered finish. The design is modern yet durable, blending with urban park aesthetics."

Drinking Fountain,"A cylindrical stainless-steel drinking fountain standing 3.5 feet tall, featuring a rounded top with a slightly recessed basin and a single arched spout operated by a side push button. The base flares subtly for stability, and the surface is brushed to reduce glare while adding texture. An additional lower basin, 2 feet high, is integrated into the side for children or wheelchair users. A discreet drainage grate encircles the base to manage overflow. The minimal, functional design complements both natural and built park elements."

Bike Rack,"A set of six inverted-U style bike racks made from powder-coated steel in matte black, each measuring 3 feet tall and 2.5 feet wide. The frames have rounded edges and a continuous tubular form for durability and ease of locking bicycles. All racks are bolted to a flat rectangular steel base plate for secure installation. The design is simple, practical, and cohesive with common park and street furniture aesthetics."

Public Art Sculpture,"A contemporary abstract sculpture composed of intersecting curved panels of weathered corten steel, forming a dynamic wave-like profile. The sculpture measures about 5 feet in length, 3 feet in width, and 3 feet in height. Its surfaces feature a rich, rust-orange patina with subtle variations in tone, giving it a warm, earthy character. The form is designed to be visually engaging from all angles, with smooth, flowing curves and cut-out sections that create shifting light and shadow patterns."

Planter with Seasonal Flowers,"A large rectangular concrete planter measuring 6 feet long, 2.5 feet wide, and 2.5 feet high, with slightly beveled top edges for a refined finish. The exterior surface is textured light-gray concrete, harmonizing with urban pavement and seating. Inside, a layered arrangement of seasonal flowers—such as bright marigolds, deep purple petunias, and tall ornamental grasses—creates a varied and colorful display. The planting design ensures year-round interest, with a mix of heights, colors, and textures for visual richness."

\begin{figure}
    \centering
    \includegraphics[width=1\linewidth]{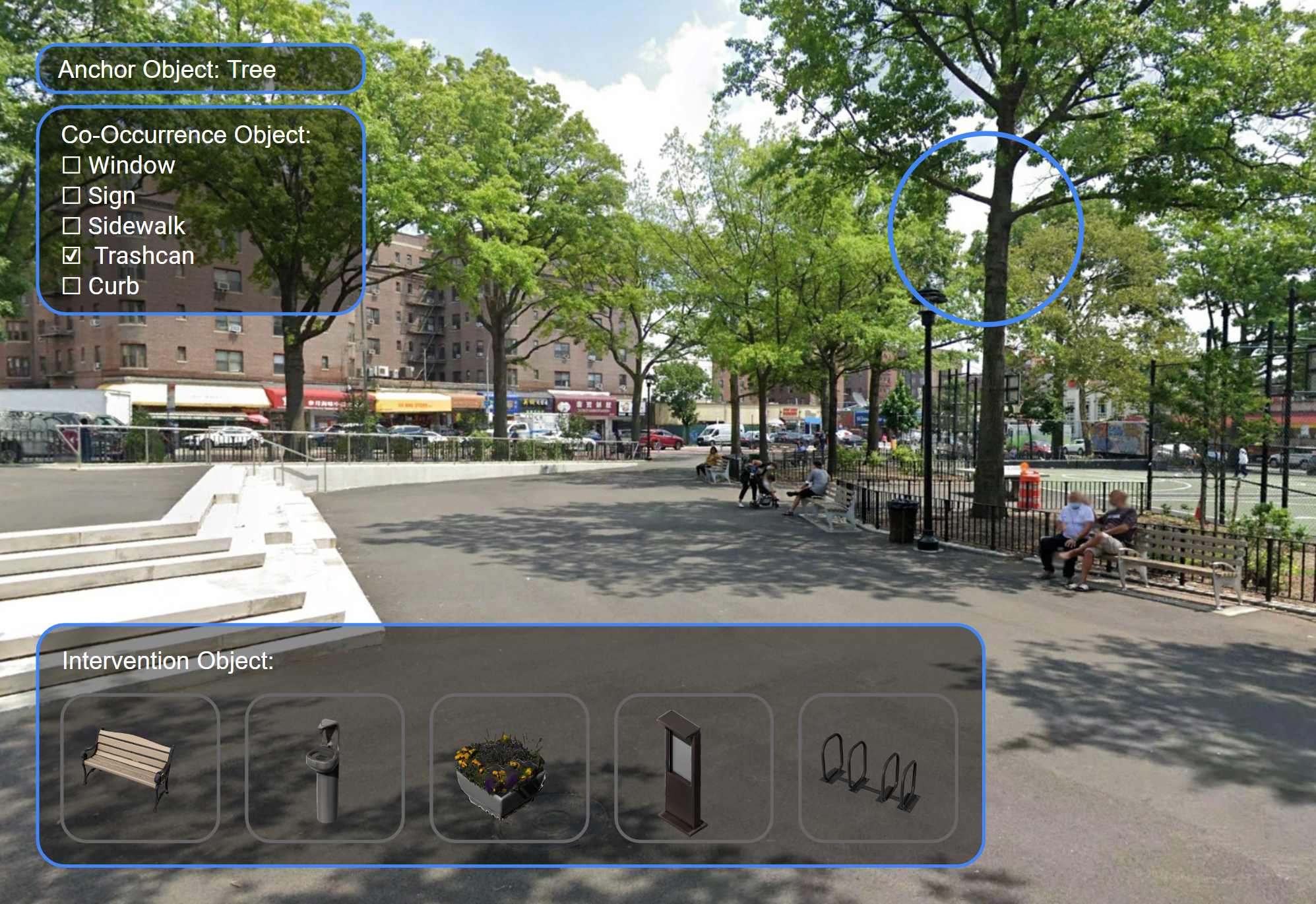}
    \caption{Pilot Interface in Urban Scene 3}
    \label{fig:placeholder}
\end{figure}

Park Bench,"A classic slatted park bench made from light beige painted wood with a black powder-coated steel frame. The bench measures approximately 6 feet in length and 3 feet in height, with a gently curved backrest and armrests at both ends for comfort. The wooden slats are evenly spaced and smooth-finished, offering durability and weather resistance. The steel frame has rounded legs with flat feet for stable ground contact. The color palette of muted beige and black harmonizes with both natural greenery and urban surroundings."

Drinking Fountain,"A modern drinking fountain crafted from brushed stainless steel, standing 3.5 feet tall with a cylindrical base and a slightly flared bottom for stability. The top has a shallow, concave basin with a chrome water spout activated by a push button. An additional side-mounted spout at 2 feet high allows accessibility for children or wheelchair users. The brushed finish reduces glare and resists fingerprints, while a circular drainage grate at the base prevents water pooling."

Bike Rack,"A modular bike rack system made from matte black powder-coated steel, featuring five inverted-U hoops aligned in a row. Each hoop stands 3 feet tall and 2.5 feet wide, made from a continuous tubular frame with smooth bends. The hoops are mounted on a rectangular steel base plate with concealed anchor bolts. The design is minimal and functional, complementing typical park fixtures and providing secure locking points for bicycles."

Information Kiosk,"A freestanding information kiosk with a rectangular frame made from dark bronze-finished aluminum, standing 6.5 feet tall and 3.5 feet wide. The top section features a clear, weather-resistant acrylic display case for maps, schedules, or community notices, with an internal LED strip for illumination. The lower section has a lockable storage compartment for maintenance tools or brochures. The overall design is sleek, durable, and styled to integrate with urban park elements."

Planter with Seasonal Flowers,"A large square planter measuring 4 feet on each side and 2.5 feet high, made from textured light-gray concrete with chamfered edges. Inside, a mix of seasonal plants such as vibrant yellow marigolds, deep purple petunias, and tall ornamental grasses create a layered, colorful display. The concrete’s neutral tone complements surrounding elements, while the plant arrangement provides year-round visual interest with varied heights and textures."

\begin{figure}
    \centering
    \includegraphics[width=1\linewidth]{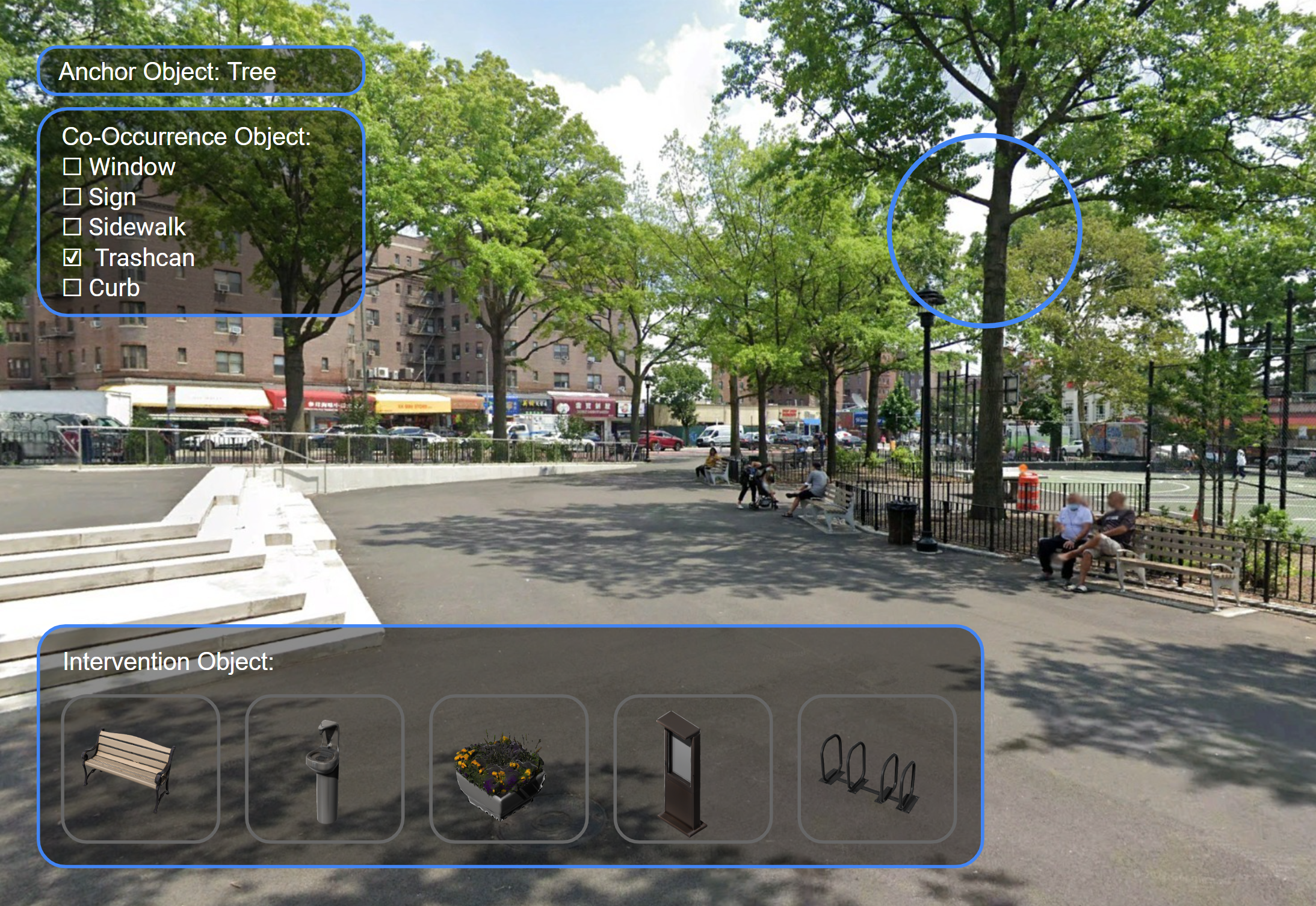}
    \caption{Pilot Interface in Urban Scene 4}
    \label{fig:placeholder}
\end{figure}

Public Seating Bench,"A contemporary public bench with a frame constructed from galvanized steel and seating surfaces made of sustainably sourced hardwood slats. The bench measures 6 feet in length, 2 feet in depth, and 3 feet in height, featuring a gently angled backrest for ergonomic comfort. The steel frame has a matte silver finish to complement surrounding urban fixtures, while the warm-toned wood slats contrast subtly with the cool pavement. The bench design incorporates slightly curved armrests and flat steel legs with base plates for stability and durability in high-traffic public spaces."

Bicycle Rack,"A modular bicycle rack made of black powder-coated tubular steel, designed in a wave pattern for both aesthetic appeal and multiple locking points. Each wave loop stands 3 feet tall and extends 2 feet in width, with smooth, continuous curves to prevent damage to bicycles. The rack is mounted on a heavy-duty steel base rail with concealed anchor bolts, ensuring secure installation. Its clean, minimalist style and durable materials match the modern urban streetscape and resist weathering from sun and rain."

Drinking Fountain,"A sleek stainless-steel drinking fountain standing 3.5 feet high, with a cylindrical pedestal base that tapers slightly toward the top. The fountain head has a push-button spout at a comfortable angle for adults, with a secondary, lower-mounted spout 2 feet high for children and wheelchair users. The brushed steel finish reduces glare and hides fingerprints, while a small grated drain surrounds the base to prevent puddling. The streamlined design fits seamlessly with contemporary city infrastructure."

Public Art Sculpture,"A small-scale abstract public art piece made from powder-coated aluminum panels in interlocking geometric shapes. The sculpture measures approximately 4 feet in height and 3 feet in width, with bold color accents in muted red, golden yellow, and steel gray to complement the tones of surrounding buildings and fixtures. The surfaces are smooth with a matte finish, and the structure is anchored to a flat steel base plate. Its compact size and modern aesthetic make it both a focal point and an unobtrusive addition to a busy streetscape."

Wayfinding Signpost,"A freestanding wayfinding signpost made from dark bronze anodized aluminum with a rectangular column design. Standing 8 feet tall and 1.5 feet wide, it features high-contrast printed panels displaying directional arrows, maps, and neighborhood points of interest. The top section includes an LED-lit panel for nighttime visibility, while the base is reinforced with a stainless-steel kick plate for durability. The clean lines and muted color palette ensure the signpost integrates harmoniously into an urban pedestrian environment."

\begin{figure}
    \centering
    \includegraphics[width=1\linewidth]{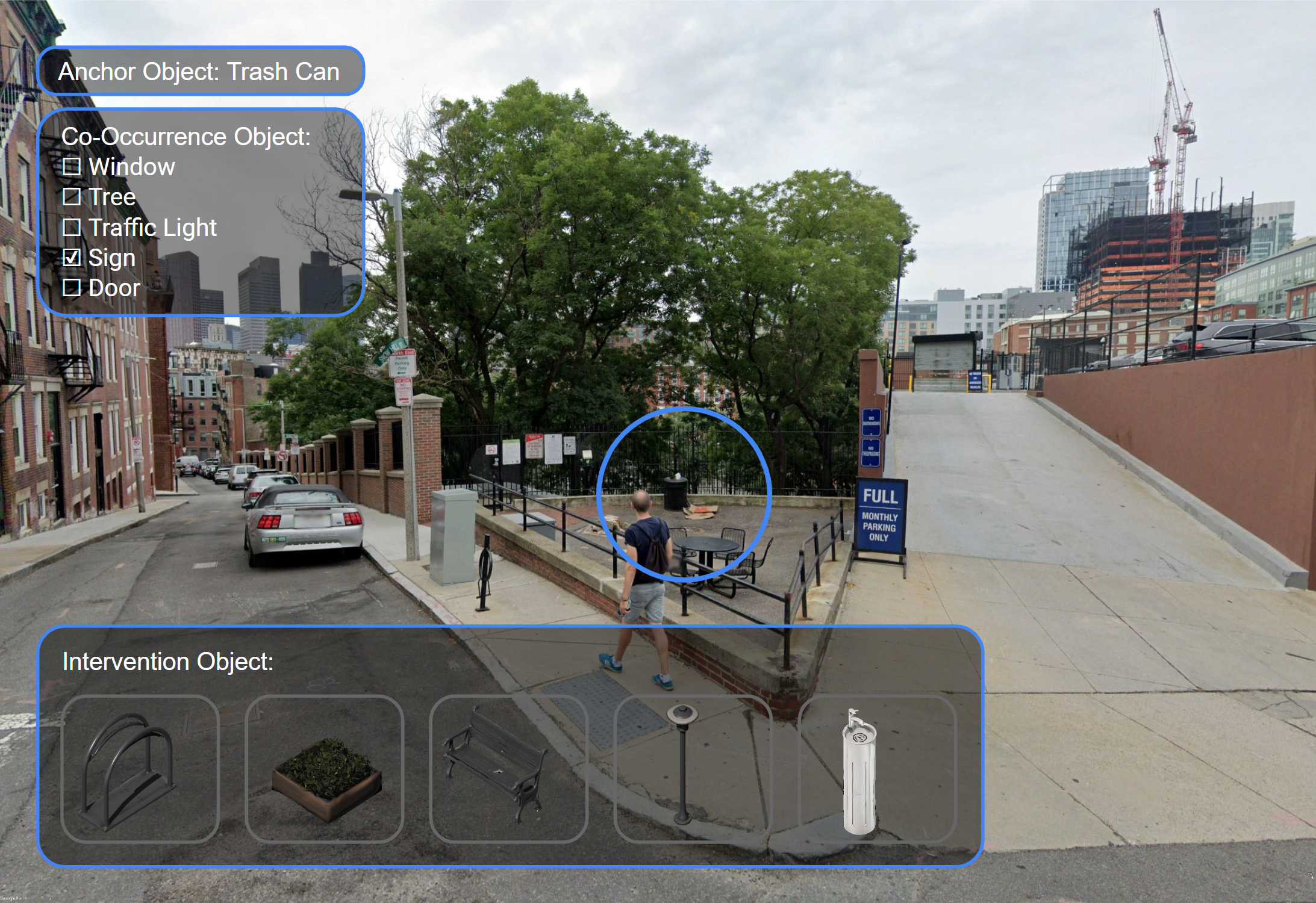}
    \caption{Pilot Interface in Urban Scene 5}
    \label{fig:placeholder}
\end{figure}

Bicycle Rack,"A black powder-coated steel bicycle rack with a simple inverted U-shape design, suitable for securing up to two bikes. The rack is made from round steel tubing approximately 5 cm in diameter, bent into a smooth arch with two vertical legs fixed to a rectangular steel base plate. The finish is matte to reduce glare, and the height is about 80 cm with a width of 90 cm, proportionate to pedestrian-scale street furniture. The style matches the utilitarian aesthetic of the surrounding urban environment, with clean lines and durability for outdoor use."

Planter Box,"A rectangular planter box constructed from weathered dark-brown cedar wood planks reinforced with black steel corner brackets. Measuring approximately 120 cm in length, 45 cm in width, and 50 cm in height, it contains low-maintenance greenery such as ornamental grasses and small flowering plants in muted greens and pale yellows that blend with the scene’s natural tones. The base has concealed drainage slots, and the top edges are slightly chamfered for a finished look, integrating natural texture into the urban setting."

Public Bench,"A three-seat metal bench with a powder-coated black steel frame and horizontal slats forming the seat and backrest. The bench is 180 cm long, 60 cm deep, and 80 cm high, with armrests at each end for comfort and accessibility. The slats are slightly curved for ergonomic support, and the style mirrors the robust, low-maintenance design typical of city street furniture. The matte finish complements the dark metal tones present in the surrounding railings and trash receptacle."

Street Lamp,"A tall, slender street lamp post approximately 3.5 meters in height, made from galvanized steel painted in a dark matte gray finish to match nearby metalwork. The lamp head features a slightly flared conical shade with a frosted glass diffuser, providing warm white LED lighting. The post tapers gently toward the top, with a simple base plate and bolted flanges for stability. Its minimalistic, functional design blends with the mix of historic brick facades and modern urban infrastructure."

Drinking Fountain,"A pedestal-style drinking fountain made from cast stainless steel with a brushed finish to resist fingerprints and weathering. The column stands about 95 cm tall with a round basin at the top and a polished chrome push-button spout. A small drainage grate is integrated into the basin to prevent standing water. The cylindrical body is sleek with subtle vertical grooves for visual interest, harmonizing with the modern utilitarian character of nearby street furniture while offering a practical amenity."

\begin{figure}
    \centering
    \includegraphics[width=1\linewidth]{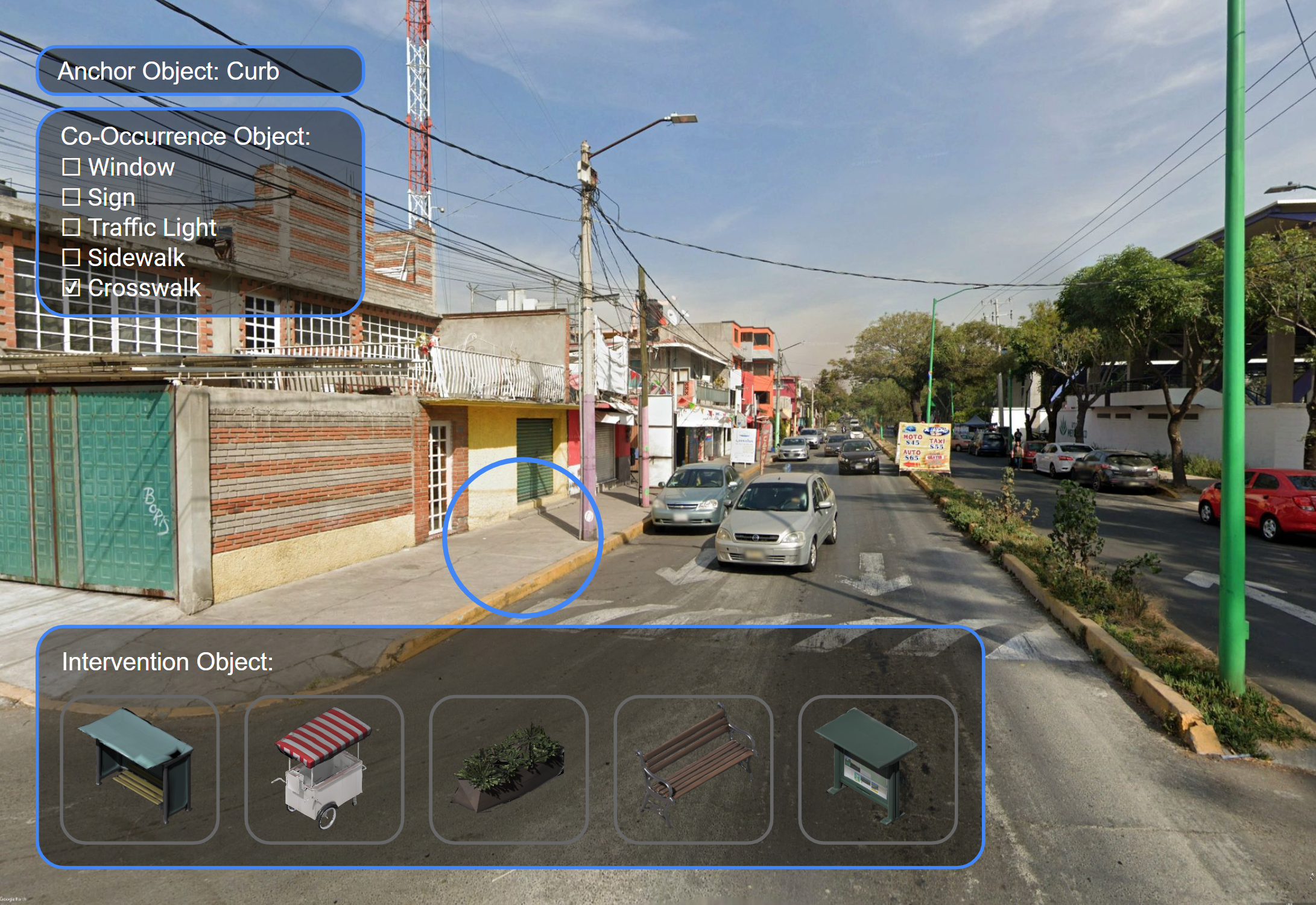}
    \caption{Pilot Interface in Urban Scene 6}
    \label{fig:placeholder}
\end{figure}

Bus Stop Shelter,"A rectangular bus stop shelter with a powder-coated steel frame in dark gray, featuring three vertical side panels made from tempered glass with a slight greenish tint. The roof is flat and slightly overhangs, constructed from opaque polycarbonate sheets to provide shade while allowing diffused light. The shelter measures approximately 3 meters in length, 1 meter in depth, and 2.4 meters in height. Inside, a single wooden bench with metal legs is mounted along the back panel. The design is minimalistic yet functional, blending with the urban streetscape mixed materials of concrete, asphalt, and painted metal."

Street Vendor Cart,"A mobile street vendor cart constructed from stainless steel with a polished surface, equipped with a canopy in red-and-white striped fabric for shade. The cart measures about 1.8 meters in length, 0.8 meters in width, and 1.5 meters in height to the canopy top. It includes side shelves for displaying goods, a small storage compartment beneath the main counter, and rubberized wheels for mobility. The style is vibrant and practical, complementing the colorful facades and signage in the surrounding area."

Planter Barrier,"A rectangular concrete planter serving as a traffic barrier, finished in a light gray tone with a slightly rough surface texture. The planter is approximately 1.2 meters long, 0.5 meters wide, and 0.6 meters tall. It contains hardy green shrubs and small flowering plants in yellow and white, adding natural contrast to the asphalt and concrete. Its utilitarian form integrates greenery while doubling as a subtle protective element for pedestrians."

Public Bench,"A metal-framed public bench with horizontal wooden slats stained in a medium brown tone, designed to seat up to three people. The bench is about 1.8 meters long, 0.6 meters deep, and 0.8 meters high, with armrests at each end for comfort. The frame is powder-coated in dark gray, echoing the muted tones of street infrastructure. The slats are slightly curved for ergonomic support, providing both aesthetic appeal and practical durability in an outdoor urban environment."

Information Kiosk,"A freestanding rectangular kiosk made of powder-coated aluminum in dark green, with a lockable glass front panel displaying local maps, transit schedules, and event posters. The kiosk measures approximately 1.5 meters in height, 0.8 meters in width, and 0.3 meters in depth. The base has a reinforced steel plate for stability, and the top features a small overhanging canopy to protect the display from rain and direct sunlight. Its color and proportions are consistent with other street fixtures, making it blend naturally into the urban scene."

\newpage
\subsection{VLM Suggestions}

%%%%%%%%%%%%%%%%%%%%%%%%%% LARGE Table test

\input{vlm_top5_stacked_all.tex}

%%%%%%%%%%%%%%%%%%%%%%%%%%% LARGE Table test

%%%%%%%%%%%%%%%%%%%%%%%%%%%%%%%%%%%%%%%%%%%%%%%%%%%%%%%%%%%%

\newpage
\section*{NeurIPS Paper Checklist}

\begin{enumerate}

\item {\bf Claims}
    \item[] Question: Do the main claims made in the abstract and introduction accurately reflect the paper's contributions and scope?
    \item[] Answer: \answerYes{}
    \item[] Justification: The abstract and introduction (Sections 1 and Abstract) clearly state the contribution, and the claims match the methods, experiments, and scope of the work.

\item {\bf Limitations}
    \item[] Question: Does the paper discuss the limitations of the work performed by the authors?
    \item[] Answer: \answerYes{}
    \item[] Justification: Section 5 discusses dataset bias, 2D-based co-occurrence limitations, and hardware accessibility constraints, along with possible future improvements.

\item {\bf Theory assumptions and proofs}
    \item[] Question: For each theoretical result, does the paper provide the full set of assumptions and a complete (and correct) proof?
    \item[] Answer: \answerNA{}
    \item[] Justification: The paper does not contain formal theorems or proofs, focusing instead on methodological and experimental contributions.

\item {\bf Experimental result reproducibility}
    \item[] Question: Does the paper fully disclose all the information needed to reproduce the main experimental results of the paper to the extent that it affects the main claims and/or conclusions of the paper (regardless of whether the code and data are provided or not)?
    \item[] Answer: \answerYes{}
    \item[] Justification: Section 3 and Appendix X describe dataset filtering, object detection settings, co-occurrence computation, and AR integration in sufficient detail for reproduction.

\item {\bf Open access to data and code}
    \item[] Question: Does the paper provide open access to the data and code, with sufficient instructions to faithfully reproduce the main experimental results, as described in supplemental material?
    \item[] Answer: \answerYes{}
    \item[] Justification: Anonymized code and data preprocessing scripts are included in the supplemental material with instructions for reproducing the main results.

\item {\bf Experimental setting/details}
    \item[] Question: Does the paper specify all the training and test details (e.g., data splits, hyperparameters, how they were chosen, type of optimizer, etc.) necessary to understand the results?
    \item[] Answer: \answerYes{}
    \item[] Justification: Dataset splits, model settings, confidence thresholds, and prompt details are described in Section 3 and Appendix X.

\item {\bf Experiment statistical significance}
    \item[] Question: Does the paper report error bars suitably and correctly defined or other appropriate information about the statistical significance of the experiments?
    \item[] Answer: \answerYes{}
    \item[] Justification: Results include mean $\,\pm\,$ standard deviation over multiple runs, capturing variability from different scene samples and random seeds.

\item {\bf Experiments compute resources}
    \item[] Question: For each experiment, does the paper provide sufficient information on the computer resources (type of compute workers, memory, time of execution) needed to reproduce the experiments?
    \item[] Answer: \answerYes{}
    \item[] Justification: Appendix X specifies GPU type, memory, and average runtime for object detection, embedding computation, and AR preview rendering.

\item {\bf Code of ethics}
    \item[] Question: Does the research conducted in the paper conform, in every respect, with the NeurIPS Code of Ethics \url{https://neurips.cc/public/EthicsGuidelines}?
    \item[] Answer: \answerYes{}
    \item[] Justification: The work follows the NeurIPS Code of Ethics, with transparency in dataset use, acknowledgment of bias, and discussion of societal impacts.

\item {\bf Broader impacts}
    \item[] Question: Does the paper discuss both potential positive societal impacts and negative societal impacts of the work performed?
    \item[] Answer: \answerYes{}
    \item[] Justification: Section X discusses benefits for community-driven design and risks such as reinforcing biased associations or privacy concerns from video capture.

\item {\bf Safeguards}
    \item[] Question: Does the paper describe safeguards that have been put in place for responsible release of data or models that have a high risk for misuse (e.g., pretrained language models, image generators, or scraped datasets)?
    \item[] Answer: \answerNA{}
    \item[] Justification: The system does not release high-risk generative models or datasets and uses only publicly available datasets with controlled prompts.

\item {\bf Licenses for existing assets}
    \item[] Question: Are the creators or original owners of assets (e.g., code, data, models), used in the paper, properly credited and are the license and terms of use explicitly mentioned and properly respected?
    \item[] Answer: \answerYes{}
    \item[] Justification: All datasets and models, including ADE20K and Grounding DINO, are cited with their licenses and source URLs.

\item {\bf New assets}
    \item[] Question: Are new assets introduced in the paper well documented and is the documentation provided alongside the assets?
    \item[] Answer: \answerNA{}
    \item[] Justification: The paper does not release a new dataset or model.

\item {\bf Crowdsourcing and research with human subjects}
    \item[] Question: For crowdsourcing experiments and research with human subjects, does the paper include the full text of instructions given to participants and screenshots, if applicable, as well as details about compensation?
    \item[] Answer: \answerNA{}
    \item[] Justification: The paper does not involve crowdsourcing or human subjects.

\item {\bf Institutional review board (IRB) approvals or equivalent for research with human subjects}
    \item[] Question: Does the paper describe potential risks incurred by study participants, whether such risks were disclosed to the subjects, and whether Institutional Review Board (IRB) approvals (or an equivalent approval/review based on the requirements of your country or institution) were obtained?
    \item[] Answer: \answerNA{}
    \item[] Justification: The paper does not involve human subjects research requiring IRB approval.

\item {\bf Declaration of LLM usage}
    \item[] Question: Does the paper describe the usage of LLMs if it is an important, original, or non-standard component of the core methods in this research? Note that if the LLM is used only for writing, editing, or formatting purposes and does not impact the core methodology, scientific rigorousness, or originality of the research, declaration is not required.
    \item[] Answer: \answerYes{}
    \item[] Justification: LLM use in the semantic branch for contextual object recommendations is described in Section 3.1 as part of the core methodology.

\end{enumerate}

\end{document}

%% file: vlm_top5_stacked_all.tex
\small
\setlength{\tabcolsep}{3pt}           % tighter inter-column spacing
\renewcommand{\arraystretch}{1.05}

% Center longtable horizontally
\setlength{\LTleft}{\fill}
\setlength{\LTright}{\fill}

% Wider first two columns, smaller right column (sums ≈ 0.98\textwidth)
\begin{longtable}{p{0.36\textwidth}p{0.36\textwidth}p{0.26\textwidth}}
\caption{VLM Top 5 (Stacked) for All Anchors}\label{tab:vlm_top5_stacked_all}\\
\toprule
\textbf{Anchor Object} & \textbf{Embedding Object} & \textbf{VLM Third Objects}\\
\midrule
\endfirsthead
\toprule
\textbf{Anchor Object} & \textbf{Embedding Object} & \textbf{VLM Third Objects}\\
\midrule
\endhead
\bottomrule
\endfoot

balcony & door & crosswalk\\
 &  & hydrant\\
 &  & bollard\\
 &  & fence\\
 &  & awning\\
\addlinespace[0.4em]
balcony & fence & traffic light\\
 &  & kiosk\\
 &  & bicycle rack\\
 &  & info board\\
 &  & curb\\
\addlinespace[0.4em]
balcony & planter & bicycle rack\\
 &  & curb\\
 &  & bollard\\
 &  & bench\\
 &  & sidewalk\\
\addlinespace[0.4em]
balcony & sidewalk & curb\\
 &  & lamp\\
 &  & trash can\\
 &  & crosswalk\\
 &  & hydrant\\
\addlinespace[0.4em]
balcony & tree & window\\
 &  & sidewalk\\
 &  & door\\
 &  & crosswalk\\
 &  & curb\\
\addlinespace[0.4em]
bench & crosswalk & info board\\
 &  & lamp\\
 &  & bollard\\
 &  & awning\\
 &  & street art\\
\addlinespace[0.4em]
bench & sign & info board\\
 &  & sidewalk\\
 &  & trash can\\
 &  & street art\\
 &  & planter\\
\addlinespace[0.4em]
bench & traffic light & tree\\
 &  & lamp\\
 &  & door\\
 &  & planter\\
 &  & bike lane marking\\
\addlinespace[0.4em]
bench & tree & bike lane marking\\
 &  & bicycle rack\\
 &  & seating wall\\
 &  & stairs\\
 &  & awning\\
\addlinespace[0.4em]
bench & window & trash can\\
 &  & bicycle rack\\
 &  & curb\\
 &  & bike lane marking\\
 &  & street art\\
\addlinespace[0.4em]
bicycle & fence & door\\
 &  & bollard\\
 &  & awning\\
 &  & bike lane marking\\
 &  & curb\\
\addlinespace[0.4em]
bicycle & pole & awning\\
 &  & kiosk\\
 &  & bollard\\
 &  & crosswalk\\
 &  & bicycle rack\\
\addlinespace[0.4em]
bicycle & sidewalk & hydrant\\
 &  & fence\\
 &  & crosswalk\\
 &  & tree\\
 &  & seating wall\\
\addlinespace[0.4em]
bicycle & traffic light & crosswalk\\
 &  & sidewalk\\
 &  & awning\\
 &  & lamp\\
 &  & stairs\\
\addlinespace[0.4em]
bicycle & window & bicycle rack\\
 &  & hydrant\\
 &  & bollard\\
 &  & planter\\
 &  & fence\\
\addlinespace[0.4em]
crosswalk & sidewalk & stairs\\
 &  & traffic light\\
 &  & bicycle rack\\
 &  & door\\
 &  & trash can\\
\addlinespace[0.4em]
crosswalk & sign & planter\\
 &  & lamp\\
 &  & curb\\
 &  & stairs\\
 &  & fence\\
\addlinespace[0.4em]
crosswalk & traffic light & trash can\\
 &  & window\\
 &  & sidewalk\\
 &  & bicycle rack\\
 &  & fence\\
\addlinespace[0.4em]
crosswalk & tree & fence\\
 &  & seating wall\\
 &  & bench\\
 &  & traffic light\\
 &  & awning\\
\addlinespace[0.4em]
crosswalk & window & tree\\
 &  & bench\\
 &  & sign\\
 &  & trash can\\
 &  & curb\\
\addlinespace[0.4em]
curb & crosswalk & planter\\
 &  & awning\\
 &  & door\\
 &  & bollard\\
 &  & fence\\
\addlinespace[0.4em]
curb & sidewalk & traffic light\\
 &  & kiosk\\
 &  & lamp\\
 &  & hydrant\\
 &  & street art\\
\addlinespace[0.4em]
curb & sign & door\\
 &  & seating wall\\
 &  & crosswalk\\
 &  & window\\
 &  & lamp\\
\addlinespace[0.4em]
curb & traffic light & seating wall\\
 &  & sidewalk\\
 &  & fence\\
 &  & awning\\
 &  & crosswalk\\
\addlinespace[0.4em]
curb & window & planter\\
 &  & bench\\
 &  & bollard\\
 &  & bicycle rack\\
 &  & street art\\
\addlinespace[0.4em]
door & crosswalk & kiosk\\
 &  & awning\\
 &  & planter\\
 &  & bollard\\
 &  & sign\\
\addlinespace[0.4em]
door & sidewalk & bench\\
 &  & tree\\
 &  & crosswalk\\
 &  & curb\\
 &  & bike lane marking\\
\addlinespace[0.4em]
door & traffic light & planter\\
 &  & crosswalk\\
 &  & sign\\
 &  & trash can\\
 &  & bollard\\
\addlinespace[0.4em]
door & tree & window\\
 &  & awning\\
 &  & planter\\
 &  & bench\\
 &  & street art\\
\addlinespace[0.4em]
door & window & awning\\
 &  & sign\\
 &  & curb\\
 &  & bollard\\
 &  & sidewalk\\
\addlinespace[0.4em]
fence & planter & curb\\
 &  & street art\\
 &  & bike lane marking\\
 &  & lamp\\
 &  & info board\\
\addlinespace[0.4em]
fence & sidewalk & hydrant\\
 &  & sign\\
 &  & trash can\\
 &  & street art\\
 &  & door\\
\addlinespace[0.4em]
fence & traffic light & window\\
 &  & street art\\
 &  & awning\\
 &  & crosswalk\\
 &  & info board\\
\addlinespace[0.4em]
fence & tree & curb\\
 &  & bike lane marking\\
 &  & traffic light\\
 &  & trash can\\
 &  & door\\
\addlinespace[0.4em]
fence & window & bench\\
 &  & bicycle rack\\
 &  & door\\
 &  & lamp\\
 &  & info board\\
\addlinespace[0.4em]
lamp & door & traffic light\\
 &  & kiosk\\
 &  & tree\\
 &  & window\\
 &  & info board\\
\addlinespace[0.4em]
lamp & sidewalk & curb\\
 &  & awning\\
 &  & hydrant\\
 &  & stairs\\
 &  & bollard\\
\addlinespace[0.4em]
lamp & stairs & bollard\\
 &  & seating wall\\
 &  & kiosk\\
 &  & street art\\
 &  & fence\\
\addlinespace[0.4em]
lamp & tree & sign\\
 &  & stairs\\
 &  & trash can\\
 &  & bollard\\
 &  & kiosk\\
\addlinespace[0.4em]
lamp & window & bike lane marking\\
 &  & info board\\
 &  & traffic light\\
 &  & planter\\
 &  & tree\\
\addlinespace[0.4em]
planter & balcony & street art\\
 &  & trash can\\
 &  & tree\\
 &  & fence\\
 &  & lamp\\
\addlinespace[0.4em]
planter & sidewalk & hydrant\\
 &  & crosswalk\\
 &  & street art\\
 &  & kiosk\\
 &  & traffic light\\
\addlinespace[0.4em]
planter & traffic light & door\\
 &  & awning\\
 &  & fence\\
 &  & trash can\\
 &  & stairs\\
\addlinespace[0.4em]
planter & tree & fence\\
 &  & trash can\\
 &  & info board\\
 &  & sign\\
 &  & window\\
\addlinespace[0.4em]
planter & window & fence\\
 &  & bench\\
 &  & bike lane marking\\
 &  & traffic light\\
 &  & lamp\\
\addlinespace[0.4em]
pole & crosswalk & sidewalk\\
 &  & bench\\
 &  & curb\\
 &  & bike lane marking\\
 &  & kiosk\\
\addlinespace[0.4em]
pole & sign & stairs\\
 &  & curb\\
 &  & info board\\
 &  & hydrant\\
 &  & seating wall\\
\addlinespace[0.4em]
pole & traffic light & planter\\
 &  & bike lane marking\\
 &  & bicycle rack\\
 &  & hydrant\\
 &  & tree\\
\addlinespace[0.4em]
pole & tree & trash can\\
 &  & traffic light\\
 &  & window\\
 &  & bollard\\
 &  & fence\\
\addlinespace[0.4em]
pole & window & awning\\
 &  & bicycle rack\\
 &  & stairs\\
 &  & bollard\\
 &  & sidewalk\\
\addlinespace[0.4em]
railing & bicycle & bollard\\
 &  & traffic light\\
 &  & hydrant\\
 &  & fence\\
 &  & stairs\\
\addlinespace[0.4em]
railing & fence & window\\
 &  & kiosk\\
 &  & info board\\
 &  & tree\\
 &  & stairs\\
\addlinespace[0.4em]
railing & pole & fence\\
 &  & bollard\\
 &  & bicycle rack\\
 &  & trash can\\
 &  & crosswalk\\
\addlinespace[0.4em]
railing & sidewalk & traffic light\\
 &  & kiosk\\
 &  & trash can\\
 &  & sign\\
 &  & seating wall\\
\addlinespace[0.4em]
railing & window & hydrant\\
 &  & street art\\
 &  & bicycle rack\\
 &  & traffic light\\
 &  & bench\\
\addlinespace[0.4em]
sidewalk & planter & crosswalk\\
 &  & tree\\
 &  & trash can\\
 &  & bike lane marking\\
 &  & street art\\
\addlinespace[0.4em]
sidewalk & sign & seating wall\\
 &  & planter\\
 &  & traffic light\\
 &  & awning\\
 &  & street art\\
\addlinespace[0.4em]
sidewalk & traffic light & awning\\
 &  & window\\
 &  & planter\\
 &  & trash can\\
 &  & kiosk\\
\addlinespace[0.4em]
sidewalk & tree & kiosk\\
 &  & crosswalk\\
 &  & window\\
 &  & traffic light\\
 &  & door\\
\addlinespace[0.4em]
sidewalk & window & curb\\
 &  & trash can\\
 &  & info board\\
 &  & street art\\
 &  & planter\\
\addlinespace[0.4em]
sign & crosswalk & trash can\\
 &  & stairs\\
 &  & bench\\
 &  & traffic light\\
 &  & window\\
\addlinespace[0.4em]
sign & sidewalk & curb\\
 &  & trash can\\
 &  & hydrant\\
 &  & planter\\
 &  & info board\\
\addlinespace[0.4em]
sign & traffic light & hydrant\\
 &  & window\\
 &  & bike lane marking\\
 &  & sidewalk\\
 &  & curb\\
\addlinespace[0.4em]
sign & tree & fence\\
 &  & bicycle rack\\
 &  & sidewalk\\
 &  & planter\\
 &  & bench\\
\addlinespace[0.4em]
sign & window & awning\\
 &  & bench\\
 &  & crosswalk\\
 &  & bike lane marking\\
 &  & bicycle rack\\
\addlinespace[0.4em]
stairs & door & fence\\
 &  & curb\\
 &  & bollard\\
 &  & bike lane marking\\
 &  & sign\\
\addlinespace[0.4em]
stairs & sidewalk & bollard\\
 &  & kiosk\\
 &  & window\\
 &  & bike lane marking\\
 &  & lamp\\
\addlinespace[0.4em]
stairs & traffic light & window\\
 &  & lamp\\
 &  & hydrant\\
 &  & tree\\
 &  & info board\\
\addlinespace[0.4em]
stairs & tree & sign\\
 &  & bollard\\
 &  & traffic light\\
 &  & window\\
 &  & fence\\
\addlinespace[0.4em]
stairs & window & bicycle rack\\
 &  & bench\\
 &  & bike lane marking\\
 &  & planter\\
 &  & kiosk\\
\addlinespace[0.4em]
traffic light & crosswalk & bike lane marking\\
 &  & awning\\
 &  & door\\
 &  & trash can\\
 &  & curb\\
\addlinespace[0.4em]
traffic light & sidewalk & planter\\
 &  & awning\\
 &  & info board\\
 &  & bike lane marking\\
 &  & tree\\
\addlinespace[0.4em]
traffic light & sign & sidewalk\\
 &  & info board\\
 &  & door\\
 &  & lamp\\
 &  & kiosk\\
\addlinespace[0.4em]
traffic light & tree & hydrant\\
 &  & sidewalk\\
 &  & bollard\\
 &  & lamp\\
 &  & bike lane marking\\
\addlinespace[0.4em]
traffic light & window & door\\
 &  & bollard\\
 &  & stairs\\
 &  & planter\\
 &  & fence\\
\addlinespace[0.4em]
trash can & door & tree\\
 &  & seating wall\\
 &  & window\\
 &  & bike lane marking\\
 &  & fence\\
\addlinespace[0.4em]
trash can & sign & crosswalk\\
 &  & tree\\
 &  & stairs\\
 &  & bike lane marking\\
 &  & kiosk\\
\addlinespace[0.4em]
trash can & traffic light & window\\
 &  & tree\\
 &  & stairs\\
 &  & crosswalk\\
 &  & street art\\
\addlinespace[0.4em]
trash can & tree & bench\\
 &  & curb\\
 &  & info board\\
 &  & kiosk\\
 &  & bike lane marking\\
\addlinespace[0.4em]
trash can & window & info board\\
 &  & tree\\
 &  & bicycle rack\\
 &  & street art\\
 &  & bollard\\
\addlinespace[0.4em]
tree & door & lamp\\
 &  & crosswalk\\
 &  & street art\\
 &  & kiosk\\
 &  & planter\\
\addlinespace[0.4em]
tree & planter & awning\\
 &  & kiosk\\
 &  & fence\\
 &  & bollard\\
 &  & sidewalk\\
\addlinespace[0.4em]
tree & sidewalk & awning\\
 &  & bike lane marking\\
 &  & kiosk\\
 &  & bench\\
 &  & bicycle rack\\
\addlinespace[0.4em]
tree & traffic light & awning\\
 &  & bicycle rack\\
 &  & stairs\\
 &  & kiosk\\
 &  & trash can\\
\addlinespace[0.4em]
tree & window & kiosk\\
 &  & stairs\\
 &  & crosswalk\\
 &  & bicycle rack\\
 &  & sign\\
\addlinespace[0.4em]
window & bicycle & sign\\
 &  & hydrant\\
 &  & crosswalk\\
 &  & awning\\
 &  & planter\\
\addlinespace[0.4em]
window & fence & bollard\\
 &  & stairs\\
 &  & street art\\
 &  & bike lane marking\\
 &  & info board\\
\addlinespace[0.4em]
window & pole & awning\\
 &  & hydrant\\
 &  & trash can\\
 &  & curb\\
 &  & traffic light\\
\addlinespace[0.4em]
window & sidewalk & lamp\\
 &  & stairs\\
 &  & traffic light\\
 &  & hydrant\\
 &  & awning\\
\addlinespace[0.4em]
window & traffic light & fence\\
 &  & trash can\\
 &  & bollard\\
 &  & door\\
 &  & bike lane marking\\
\addlinespace[0.4em]

\end{longtable}